\pdfoutput=1
\documentclass[11pt]{article}

\usepackage{ACL2023}

\usepackage{times}
\usepackage{latexsym}
\usepackage{xcolor}
\usepackage{graphicx}

\usepackage[T1]{fontenc}
\usepackage[utf8]{inputenc}

\usepackage{microtype}

\usepackage{inconsolata}
\usepackage{multirow}

\title{Hierarchy Builder: \\Organizing Textual Spans into a Hierarchy to Facilitate Navigation}

 \author{Itay Yair$^\dagger$ \and Hillel Taub-Tabib$^\ddagger$ \and Yoav Goldberg$^{\dagger,\ddagger}$ \\
 $^\dagger$Computer Science Department, Bar Ilan University \\
 $^{\ddagger}$Allen Institute for Artificial Intelligence \\
 }
\begin{document}
\maketitle
\begin{abstract}
Information extraction systems often produce hundreds to thousands of strings on a specific topic. We present a method that facilitates better consumption of these strings, in an exploratory setting in which a user wants to both get a broad overview of what's available, and a chance to dive deeper on some aspects. The system works by grouping similar items together, and arranging the remaining items into a hierarchical navigable DAG structure. We apply the method to medical information extraction.
\end{abstract}

\section{Introduction}

\begin{figure*}[t!h]
    \centering
    \includegraphics[width=0.9\linewidth]{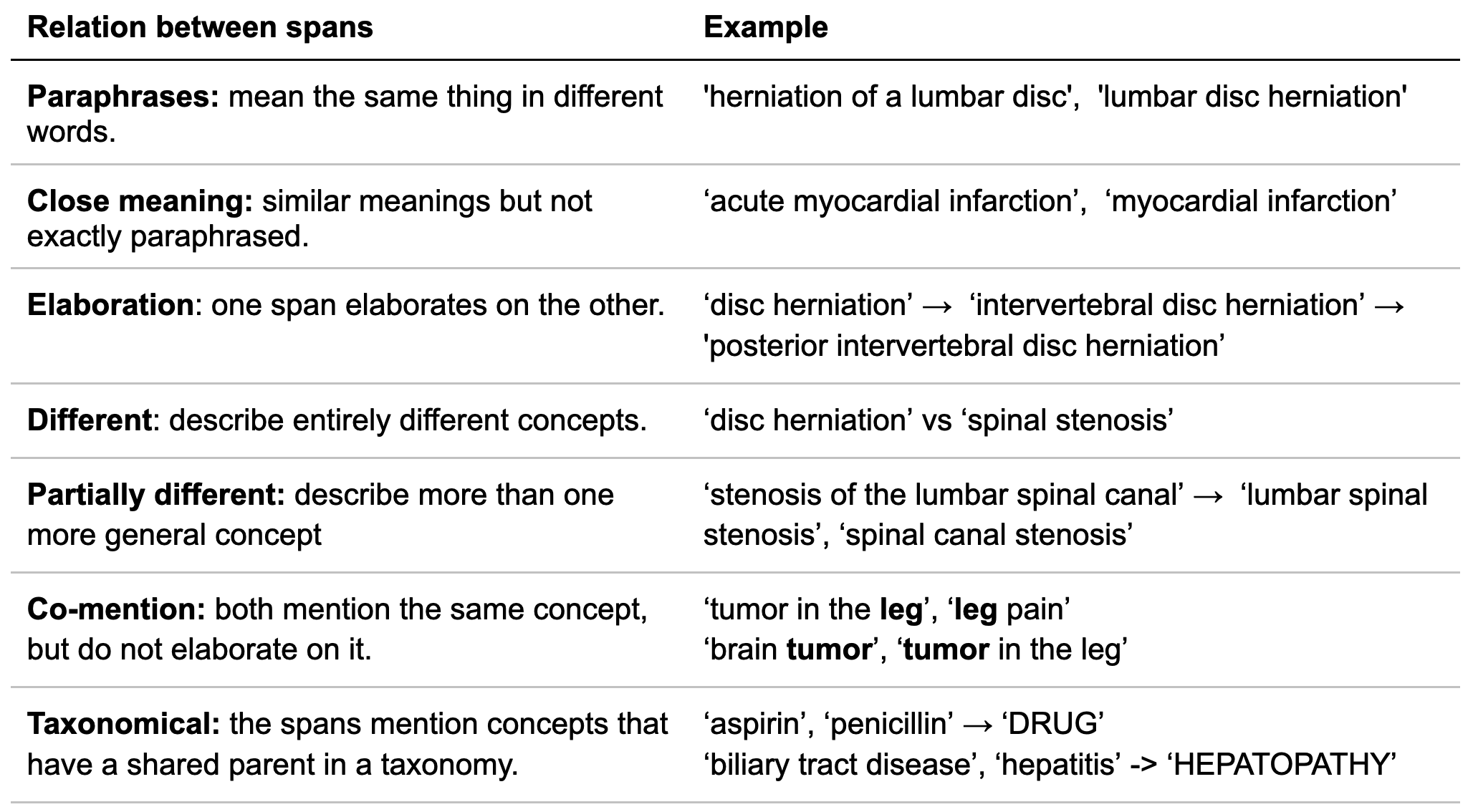}
    \caption{Kinds of possible relations between input strings}
    \label{fig:relation}
\end{figure*}

We are dealing with the question of organizing and displaying a large collection of related textual strings. The need arises, for example, in information extraction or text mining applications, that extract strings from text.
Consider a system that scans the scientific literature and extracts possible causes for a given medical condition. Such a system may extract thousands of different strings, some of them relate to each other in various ways,\footnote{Figure \ref{fig:relation} lists the kinds of relations between strings.} and some are distinct.
Users consume the list in an exploratory mode \cite{agarwal2021lookup}\cite{white2008evaluation}, in which they do not have a clear picture of what they are looking for, and would like to get an overview of the different facets in the results, as well as to dig deeper into some of them. 

For example, distinct strings extracted as causes for sciatica include ``\emph{herniated disc}'', ``\emph{herniated disk}'', ``\emph{lumbar disk herniation}'' , ``\emph{posterior interverbal disc herniation}'' and ``\emph{endometriosis}'', among hundreds of others. The user of this system likes to go over the returned list to learn about possible causes, but going over hundreds to thousands of results is mentally taxing, and we would like to reduce this effort. In the current case, we would certainly like to treat the first two items (\emph{herniated disc} and \emph{herniated disk}) as equivalent and show them as one unified entry. But we would also like to induce an additional hierarchy. For example, it could be useful to separate all the \emph{herniated disc} related items (or even all the \emph{disc} related items) in one branch, and the \emph{endometriosis} case in another. This will allow the user to more efficiently get a high level overview of the high-level represented topics (\emph{disc herniation} and \emph{endometriosis}) and to navigate the results and focus on the cases that interest them in the context of the query (for example, they may feel they know a lot about disc-related causes, and choose to ignore this branch).

An additional complication is that the hierarchy we are considering is often not a tree: a single item may have two different parents, resulting in a direct acyclic graph (DAG).  For example, arguably a condition like \emph{leg pain} should be indexed both under \emph{leg} (together with other leg related items) and under \emph{pain} (together with pain related items). The hierarchy structure is contextual, and depends on the data: if there are not many other leg related items, it may not be beneficial to introduce this category into the hierarchy.

Additionally, note that some items in the hierarchy may not directly correspond to input strings: first, for the ``\emph{leg pain}'' example above, if the input list does not include stand-alone \emph{leg} or \emph{pain} items, we may still introduce them in our hierarchy. We may also introduce additional abstraction,  for example we may want to group ``\emph{heart disease}'', ``\emph{ischemia}'', ``\emph{hypotension}'', and ``\emph{bleeding}'' under ``\emph{cardiovascular disease}''.

In this work we introduce a system that takes such a flat list of related strings, and arranges them in a navigable DAG structure, allowing users to get a high level overview as well as to navigate from general topics or concepts to more specific content by drilling down through the graph. Ideally, the graph would allow the user to:\\
(1) get a comprehensive overview of the various facets reflected in the results;\\
(2) quickly get an overview of the main aspects of the results;\\
(3) efficiently navigate the results, finding items in the sub-graph in which they expect to find them.

At a high level, the system works by finding lexically equivalent terms, arranging them in a DAG structure reflecting the specificity relation between terms, further merging equivalent nodes based on a neural similarity model, add additional potential intermediary hierarchy nodes based on taxonomic information and other heuristics, and then pruning it back into a smaller sub-DAG that contains all the initial nodes (input strings) but only a subset of the additional hierarchy nodes. Finally, we select the top-k ``entry points'' to this graph: high level nodes that span as many of the input nodes as possible.  This process is described in section \S\ref{sec:methods}. While the DAG extended with potential hierarchies is very permissive and contains a lot of potentially redundant information, the DAG pruning stage aims to ensure the final graph is as compact and informative as possible.

We focus on causes-for-medical-conditions queries and provide a demo in which a user can select a medical condition, and browse its causes in a compact DAG structure.

To evaluate the resulting DAGs, we perform automatic and manual evaluation. The automatic evaluation is based on measuring various graph metrics. The human evaluation is performed by human domain experts. Our results show that the DAG structure is significantly more informative and effective than a frequency-ranked flat list of results.

\section{Requirements}

As discussed in the introduction, our input is a list of strings that reflect answers to a particular question, as extracted for a large text collection (we focus in this paper on the biomedical domain, and more specifically in causes for medical conditions). This list can be the output of an Open-IE system \cite{fader-etal-2011-identifying,stanovsky-etal-2015-open,kolluru-etal-2020-openie6}, the results of running extractive QA \cite{rajpurkar-etal-2016-squad} with the same question over many paragraphs, or extracted using an extractive query in a system like SPIKE \cite{shlain-etal-2020-syntactic,taub-tabib-etal-2020-interactive,ravfogel-etal-2021-neural}. The lists we consider typically contain from hundreds to thousands of unique items. We identified a set of relations that can hold between strings in our inputs, which are summarized in Table \ref{fig:relation}.
We would like to arrange these items in a hierarchical structure to facilitate exploration of the result list by a user, and allow them to effectively consume the results. Concretely, the user needs to:\\
\emph{ a. not see redundant information.}\\[0.5em]
\emph{ b. be able to get a high-level overview of the various answers that reflected from the results.}\\[0.5em]
\emph{ c. be able to get a quick access to the main answers.}\\[0.5em]
\emph{ d. be able to dig-in into a specific phenomenon or concept that is of interest to them.}\\[0.5em]
\emph{ e. be able to locate concepts they suspect that exist.}

This suggests a hierarchy that respects the following conditions:\\
\emph{Paraphrased} spans should be combined into a single group, and \emph{close-meaning} spans should be combined into the same group;
\emph{Elaboration} relations should be expressed hierarchically;
\emph{Co-mention} spans should be both descendants of the shared concept;
\emph{Taxonomic relations} should (in some cases) be descendants of the taxonomical parent.

Additionally, we would like each node in the hierarchy to have relatively few children (to reduce the need to scan irrelevant items), yet keep the hierarchy relatively shallow (to save expansion clicks if possible). The hierarchical structure should also be informative: we should be able to guess from a given node which kinds of items to expect to find under it, and which kinds of items \emph{not} to expect to find under it. This means a single item should be lockable in different ways, in case it can be categorized under different keys (we would sometimes like ``\emph{brain tumor}'' to be listed under \emph{brain} and sometimes under \emph{tumors}).\footnote{Arranging information as graphs to facilitate navigation and exploration is, of course, not a novel concept. A notable examples is entailment graphs \cite{kotlerman2015textual,adler-etal-2012-entailment}.}

\section{Method}
\label{sec:methods}
\paragraph{Expanding the initial list.}
We assume that the strings in the initial list are \emph{maximal}, meaning that the string captures the extracted noun-phrase including all of its possible modifiers. We further expand the list by considering also potential substrings of each maximal string, reflecting different granularities. For example, from the string ``severe pain in the lower right leg'' we would extract ``pain'', ``severe pain'' , ``severe pain in the leg'', ``severe pain in the lower right leg'', among others.\footnote{This is done using a rules-based algorithm that operated on the parse tree, which extracted all the distinct modification spans derived from the head token.}
We then consider the union of the initial set of input strings and the set of additional sub-strings. Different users would be interested in different granularities depending on their information need. We rely on the DAG-pruning stage to properly organize these strings and prune away non-informative ones in the context of the entire set.
%We expand the initial set of strings by adding also all nouns and adjectives that appear in any of the strings, and which are not already represented as standalone items in the list. 

\paragraph{Initial grouping into equivalence sets.}
The input of this stage is a set of strings (the union of the input set and the extended set), and the output is a list of sets, such that the sets are distinct, and their union covers the initial set. For example, after this stage, the items ``\emph{herniated disk}'', ``\emph{herniated disc}'', ``\emph{disc herniation}'', ``\emph{herniation of the disc}'' will be in the same equivalence set.

The grouping in this stage is inspired by \cite{gashteovski-etal-2017-minie} and is based on considering each string as a bag of lemmas, discarding stop words, modal words, and quantity words, and considering items as equivalent if their bags are equivalent. The lemma matching is relaxed, and allows, beyond exact string match, also matches with small edit distance and matches based on UMLS \cite{10.1093/nar/gkh061} and WordNet \cite{miller-1992-wordnet} spelling variants and synonyms.

\paragraph{Initial DAG construction.}
We now take the list of sets from the previous stage, and arrange them into a DAG, where each set is a DAG node. We add a directed edge between two nodes A and B if B \emph{is more specific than} A, and no other node C is more specific than A and less specific than B.
 
The \emph{specificity relation} at this stage is determined based on the bags of lemmas that were used to create the equivalence sets: a set B is more specific than a set A if A and B are not equivalent and the bag of B contains the bag of A. 

\paragraph{Adding heads as nodes}
For all spans, we take their head-word (either a single adjective or a single noun) and add them as roots of the DAG. We then add an additional root node above them, so that the DAG has a single root. This handles the co-mention relation.

\paragraph{Merging semantically equivalent graph nodes.}

We now take the DAG and merge equivalent nodes, as determined by a trained statistical model (we use SAP-BERT \cite{liu2020self})\footnote{We chose SAP-BERT for its entity-linking specialization, and since it outperformed other models we tried, such as Sci-Bert\cite{beltagy-etal-2019-scibert}, in detecting semantic similarity for our specific case.}. For example, this stage will merge ``\emph{administration of streptozotocin}'' and ``\emph{streptozotocin injection}''. When merging two graph nodes, we handle the corresponding edges in the expected way (the children of the two individual nodes become children of the merged node, and the parents of the individual nodes become the parents of the merged node).\footnote{We perform this stage after the DAG construction and not prior to it, as it makes the specificity relation between nodes significantly harder to define. In the current order, we first define specificity based on lexical containment, and then add further merge the groups.}

For a pair of graph nodes A and B, we encode each string in A and in B into a vector using SAP-BERT, and represent each node as the average vector of the strings within it. 
We go over the nodes in the DAG in DFS order starting from the root nodes, and for each node consider all of its children for potential merging among them. We merge two nodes if the cosine similarity score between their vectors passes the threshold $t_1 = 0.9$ and their merging does not create a cycle.
We then do another pass and merge nodes to direct child nodes if their similarity score is above $t_2 = 0.95$, again avoiding creating circles.

After this stage, we attempt to further merge nodes based on the UMLS ontology \cite{10.1093/nar/gkh061}. Two nodes A and B are considered UMLS-equivalent, if there is at least one string in node A that is listed in UMLS as a synonym of at least one string in node B. Such cases are merged.\footnote{If this merging creates a cycle, this cycle is removed.}

\paragraph{Adding taxonomic nodes.}
So far the relationships between nodes in the DAG were solely based on lexical relations. In order to enrich the graph, we introduce additional nodes based on taxonomical relations, which are not reliant on lexical information. For instance, ``heart disease'', ``ischemia'', ``hypotension'', and ``bleeding'' are under the broader term ``cardiovascular disease''.
We add many nodes here, relying on many of them to be pruned in the next stage.

We map each node to the UMLS hierarchy, and look for UMLS concepts that govern at least two DAG nodes (``descendent DAG nodes''). These are potential abstractions over graph nodes. For each such UMLS concept that is already part of the DAG, it is connected by an edge to all its descendant DAG nodes that do not already have a path to them, if adding such an edge does not create a cycle. For UMLS concepts that are not already in the DAG, they are added as new nodes governing the descendant graph nodes.
UMLS concepts have multiple synonyms. When adding them as nodes, we choose the synonym with the highest SAP-BERT cosine similarity to the descendent DAG nodes this concept governs.

\paragraph{DAG Pruning.}
The DAG at this stage is quite large and messy, containing both nodes containing input strings, as well as additional hierarchy nodes based on linguistically motivated substrings of the input strings, and on taxonomic relations.  We prune it to create a smaller graph which is more amenable to navigation. The smaller DAG should contain all the nodes corresponding to input strings, and an effective set of additional hierarchy nodes. 
Some of the hierarchy nodes are more important than others, as they provide a better differential diagnosis among the answers. Our goal is to highlight these and filter out the less important ones. Operatively, we would like for each node in the graph to have the minimal number of children, such that all the input strings that were reachable from it, remain reachable from it. This focuses on hierarchy nodes that are shared among many input concepts. 

We first prune graph edges according to this criteria. This process results in nodes that have a single child. Such nodes are removed, and their children are attached to their parent.\footnote{Selecting the smallest group of concepts at each hierarchy level is important for user navigation, who quickly become overwhelmed by too many nodes, making it difficult to orient themselves within the DAG.}

Selecting the minimal number of children according to this criteria is NP-hard.
As an alternative, we use an approximation algorithm called the greedy set cover algorithm \cite{johnson1973approximation}, which works by selecting in each step the node with the highest number of non-covered answers, covering them, and proceeding. This helps in choosing the most important concepts and with the highest differential diagnosis.

\begin{figure*}[t!]
  \includegraphics[width=\textwidth]{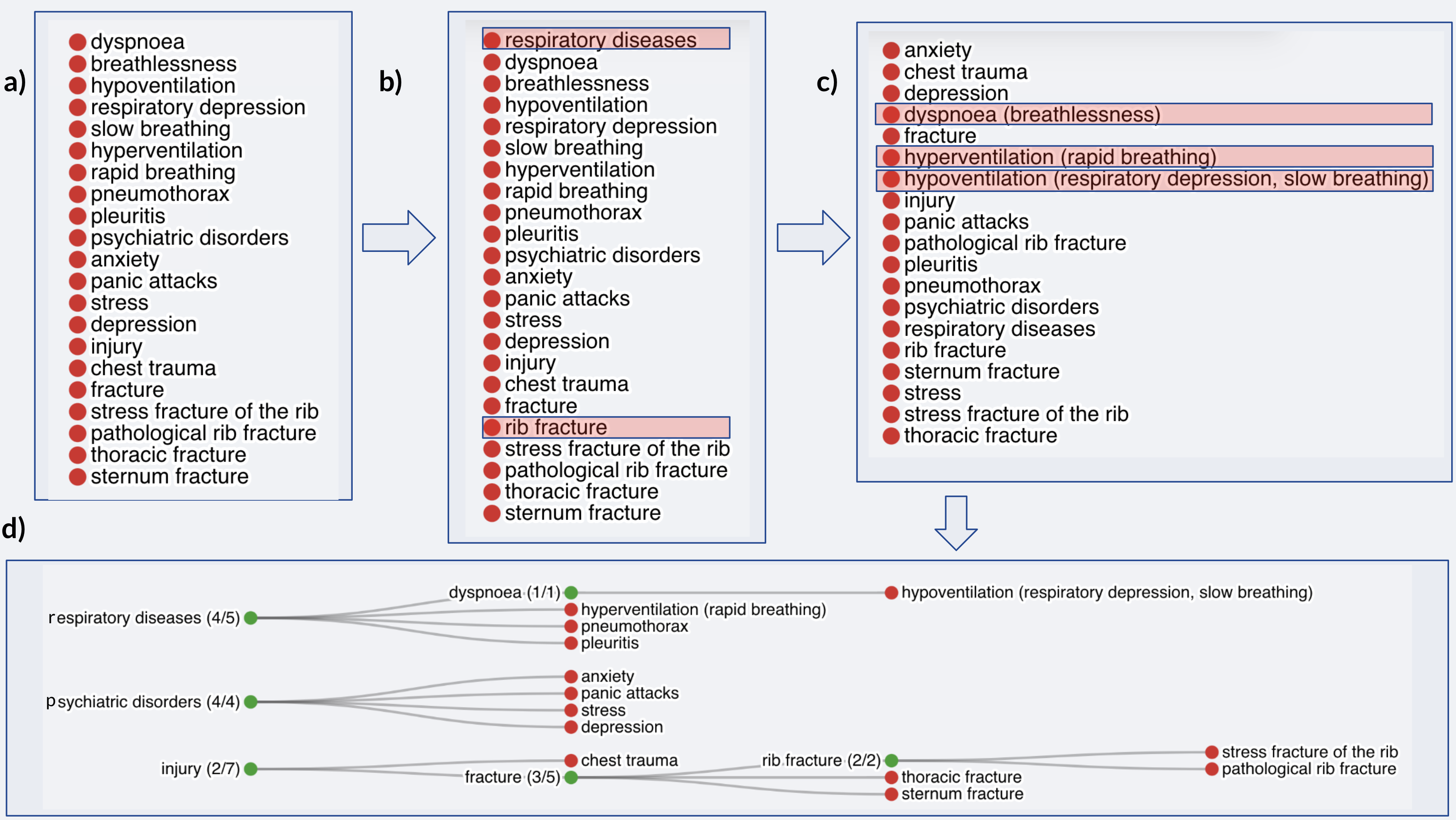}
  \caption{Input-Output Example. See section \S\ref{sec:example}.}
  \label{fig:io_example}
\end{figure*}

\paragraph{Entry-point selection.}
Finally, we seek $k$ nodes that will serve as the ``entry nodes'' to the graph. These should be $k$ nodes that fulfill the following criteria:\\
a. allow reaching as many input strings as possible.\\
b. the semantic affinity between a node and the input string reachable by it, is high.

The users will initially see these nodes as well as an additional ``other'' node, from which all the other input strings can be reached. The entry node labels provide an overview of the $k$ main concepts in the list, and allow the user to both get an overview of the result as well as to drill down into parts that interest them. Criteria (b) is important to ensure that the user not only can reach the input string by navigating from an entry point, but also that it will \emph{expect} to find this input string there.

This selection is done by a heuristic algorithm which we adapted from the Greedy+ DAG-node-selection algorithm in \cite{zhu2020top}.
It first assigns each node C with a score that combines the number of the input nodes reachable from it, and the semantic affinity (based on SAP-BERT cosine similarity) of C to each of these reachable nodes. It then iteratively adds the highest scoring candidate C to the set of entry points, and adjusts the scores of each remaining node N by subtracting from the score of N the affinity scores between C and the input nodes reachable from N. We do this until we reach $k$ entry points.

\paragraph{Visualization.} We are now finally ready to show the DAG to the user. For nodes that correspond to multiple (semantic equivalent but lexically different) input strings, we choose one of them as the representative for display purposes.

\section{Input-output Example}
\label{sec:example}
We demonstrate with a minified example.
Given the set of spans in Figure (\ref{fig:io_example}a), representing causes of chest pain, Hierarchy Builder expands the set by adding the spans "rib fracture" (this is a substring of two existing spans) and "respiratory diseases" (a new taxonomic node). Based on the expanded set of spans in Figure (\ref{fig:io_example}b), Hierarchy builder identifies synonymous spans and merges them into the concepts. In Figure (\ref{fig:io_example}c) we see these concepts, where each concept includes aliases in parenthesis where applicable. Hierarchy Builder then places the entries in a DAG based on a hierarchy of specificity, as depicted in Figure (\ref{fig:io_example}d).

\section{Experiments and Evaluation}

\paragraph{Scope}
We focus on the medical domain and evaluate our system on etiologies (causes) of two medical symptoms (``\emph{jaundice}'' and ``\emph{chest pain}''). These symptoms were chosen because their are common and each contains many different etiologies mentioned in the literature. 

The input lists for the system were the result of running a set of 33 syntactic patterns over PubMed abstracts, looking for patterns such as ``\texttt{COND due to \_\_\_}'' or ``\texttt{patients with COND after \_\_\_}'' where COND is either \emph{jaundice} or \emph{chest pain}. The results were extracted using the SPIKE system \cite{shlain-etal-2020-syntactic,taub-tabib-etal-2020-interactive} and each matched head-word was expanded to the entire syntactic subgraph below it. This resulted in 3389 overall extracted strings and 2623 unique strings for \emph{jaundice} and 2464 overall and 2037 unique for \emph{chest pain}. After merging strings into synonym sets as described in \S\ref{sec:methods}, we remain with 2227 concepts for \emph{jaundice} and 1783 for \emph{chest pain}.

For each of the symptoms there are established and widely accepted lists of common etiologies, which we rely on in our evaluation.\footnote{We take the established etiologies for \emph{jaundice} from \url{https://www.ncbi.nlm.nih.gov/books/NBK544252/} and for \emph{chest pain} from \url{https://www.webmd.com/pain-management/guide/whats-causing-my-chest-pain}.}
We take 38 established etiologies for jaundice and 33 for chest pain, and check their accessability in the flat list of extracted symptoms, as well as in the hierarchical DAG we create.

\paragraph{Coverage and Entry-point Selection}
For \emph{jaundice}, our input list contains 28 out of the 38 known etiologies, and for \emph{chest pain} 26/33. With $k=50$, 25 of 28 concepts are reachable from an entry point for \emph{jaundice} and 21/26 for \emph{chest pain}. With $k=100$ the numbers are 28/28 (\emph{jaundice}) and 24/26 (\emph{chest pain}).

\begin{table*}[t!h]
    \centering
    \begin{tabular}{p{0.4\textwidth}p{0.15\textwidth}p{0.15\textwidth}p{0.15\textwidth}}
    \textbf{Component} & \textbf{Contribution} & \textbf{Chest-pain} & \textbf{Jaundice} \\
    \hline
    Expanding the initial list (for full DAG) & Add nodes & 504 & 893 \\
    Expanding the initial list (for DAG with 50 entry nodes) & Add nodes & 158 (12~top~level) & 350 (17~top~level) \\
   \hline
    Adding heads as nodes (Full DAG) & Add nodes & 457 & 379 \\
    Adding heads as nodes (50 entry nodes) & Add nodes & 20 (5~top~level) & 19 (6~top~level) \\
    \hline
    Merging semantically equivalent nodes & Merge nodes & 93 (out of 2556) & 266 (out of 3330) \\
    UMLS merging of synonym nodes & Merge nodes & 62 (out of 2504) & 99 (out of 3167) \\
    \hline
    UMLS taxonomic nodes (full DAG) &  Add nodes & 113 & 169 \\
    UMLS taxonomic nodes (50 entry nodes) & Add nodes & 3 & 6 \\
    \hline
    UMLS taxonomic edges & Add edges & 140 (5~top~level) & 153 (3~top~level) \\
    \hline
    DAG Pruning & Remove edges & 2363 & 3209 \\
    \end{tabular}
    \caption{Quantifying the contribution of the different components.}
    \label{fig:components}
\end{table*}

\paragraph{Assessing the contribution of the different components}
The different components in our algorithm contribute by adding nodes, combining nodes, adding edges, and removing edges. 
Table \ref{fig:components} describes the kind of contribution of each component and quantifies its impact, for each of the two tested conditions. 

We now look at the case where we select 50 entry-point nodes, and focus on the effect on the top-level nodes. We see that for Chest-pain, a total of 20 of the 50 selected entry-points were not in the original input, but were added by the various components (12 from expanding the initial list, 5 from adding head words, and 3 from taxonomic words). Similarly, for Jaundice, these components added a total of 29 root nodes (out of the selected 50) that were not in the original input (17 from expanding initial list, 5 from head words and 6 from taxonomic nodes).

The ``Expanding the initial list'' component plays a significant role in shaping the DAG structure. In Chest Pain, 161 out of 224 internal nodes originate from the expanded list (146 from Expanding the initial list and 15 from co-mention). In Jaundice, 347 out of 423 internal nodes stem from the expanded list (333 from Expanding the initial list and 14 from co-mention). This highlights the substantial impact of this component on the DAG's structure.

The number of merges performed indicates the usefulness of the employed merging methods.

Furthermore, the set cover pruning algorithm effectively reduces the number of edges in the DAG.

\paragraph{Qualitative Measures}
For \emph{jaundice}, our final DAG contains 2620 nodes overall and has a maximum depth of 11. With $k=50$ The average number of leaves per entry point is 22.68 (min 0, max 600), and the average depth is 2.86 (min 0, max 9). Most importantly, each internal node has an average of 9.12 children (min 1, max 56, variance 34.91), making them highly browsable.

For \emph{chest pain}, the trends are overall similar: our final DAG contains 2124 nodes overall and has a maximum depth of 9. With $k=50$ The average number of leaves per entry point is 14.14 (min 1, max 175), and the average depth is 2.8 (min 0, max 7). Each internal node has an average of 4.94 children (min 1, max 53, variance 27.53).

\paragraph{Human evaluation.}
Our main evaluation centers around the effort for an expert\footnote{We use two experts, each evaluating a different condition. The expert evaluating \emph{jaundice} is an expert MD specializing in children's medicine. The expert evaluating \emph{chest pain} is a PhD in biology with 38 years of biomedical research.} to locate the known etiologies in the resulting DAG, compared to a flat list sorted by frequency. For each of the etiologies, we ask how many entries need to be considered before finding the etiologies. For the flat list, this means how many items are read when scanning the list in order before reaching the etiology. For the DAG, we count the number of clicks (expansions of a node) starting from $k=50$ entry points (a quantity that aligns with a reasonable threshold of entry nodes perceivable by a user)
, while summing also the number of items before the expanded node in each level. Note that since we look for common etiologies rather than rare ones, we would assume a frequency-ranked list based on literature mentions would compare favorably in these measures. Nonetheless, we see a clear benefit of the DAG. We compare to conditions: an ideal condition where the user knows exactly which nodes to expand (blue in the graph), and a realistic scenario, in which the user searches for the etiologies by expanding nodes (gray in the graph).

We also perform another evaluation in which we ask the experts to rank each path to an etiology based on its quality, given the question ``to what extent is this a logical path to follow in order to find the etiology'', on a scale of 1 (very bad) to 5 (very good). 

\begin{figure}[t!]
    \centering
    \includegraphics[width=0.49\textwidth]{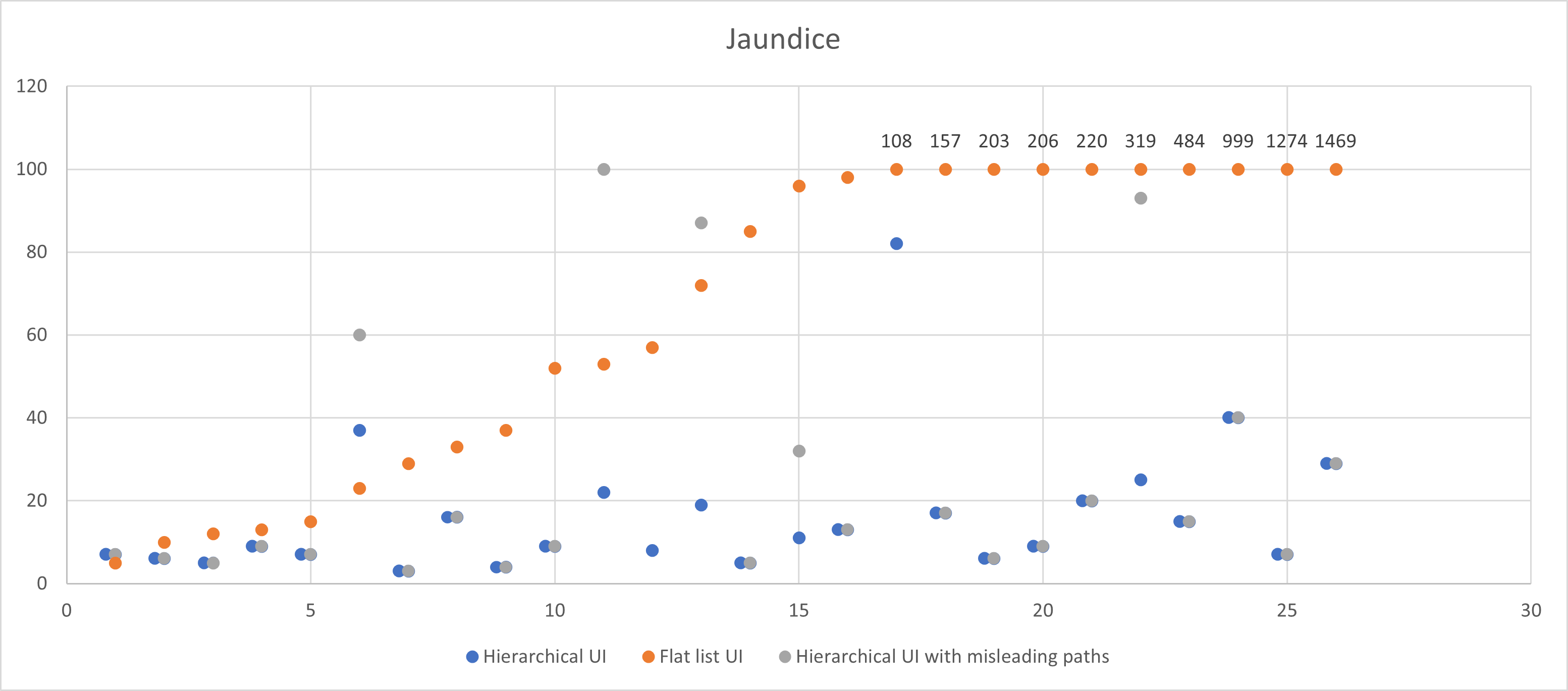} \\
    \includegraphics[width=0.49\textwidth]{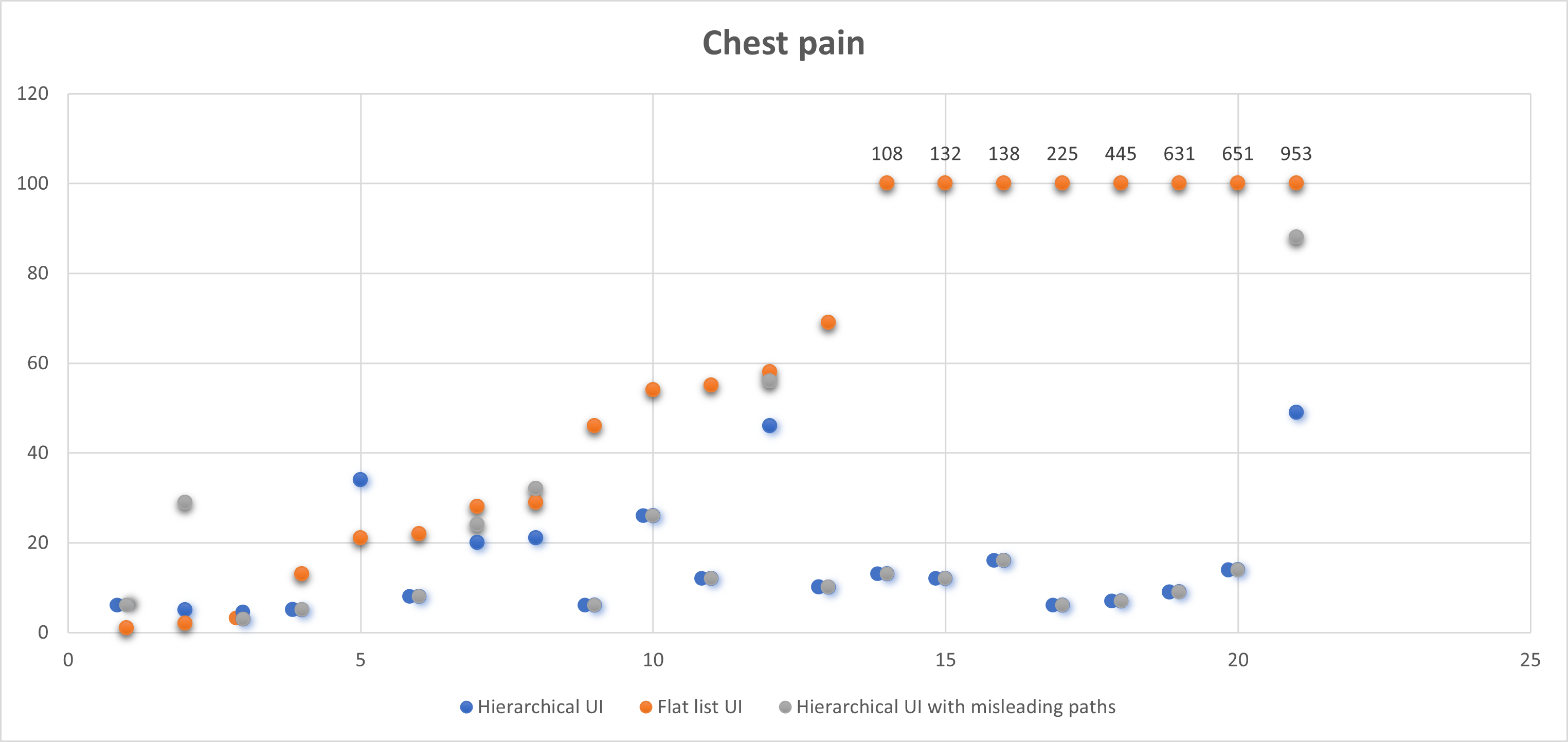}
    \caption{Effort to reach a set of common etiology items using our created DAG vs. a frequency ranked list. X axes coordinates correspond to different etiologies sorted by their frequency in the input list, and Y axes correspond to the effort. Orange: frequency-ranked flat list. Blue: DAG + oracle locating of items. Gray: DAG + human locating of items.} 
    \label{fig:res}
\end{figure}

\paragraph{Results} Figure \ref{fig:res} shows the main results for the two conditions. Despite the frequency-based ranking, many of the etiologies appear relatively low in the flat list, making them very hard to come by in this condition (orange). On the other hand, when considering the DAG, the vast majority of items are significantly easier to locate, requiring scanning significantly fewer items. Only 3 items for jaundice and 2 for chest pain were significantly harder to locate in the DAG than in the flat list. In terms of the quality of the DAG paths associated with each etiology, the jaundice annotator ranked 23 out of 25 as 5, 1 as a 2, and 1 as a 1. For chest pain, the numbers are 19 out of 21 ranked as 5, 1 as 2, and 1 as 1. Overall, our hierarchy building algorithm works well for the vast majority of the cases, and offers significant benefits over the flat list.

\section{Conclusions}
We presented an automatic method to organize large lists of extracted terms (here, of medical etiologies) into a navigable, DAG-based hierarchy, where the initial layer provides a good overview of the different facets in the data, and each internal node has relatively few items. The code together with a video and an online demonstration are available at \url{https://github.com/itayair/hierarchybuilder}.

\section{Limitations}
While our method is aimed at organizing any flat-list of extractions, we evaluated it here only on the medical domain, only on a single kind of information need (etiologies), and only for common conditions (jaundice and chest pain). More extensive evaluation over additional conditions is needed in order to establish general-purpose utility. However, we do find the system useful for navigating in automatically-extracted etiology lists, and encourage the readers to experiment with the system also on other conditions, to assess its utility.

There are also some candidates for improving the method also in the biomedical domain, which are not currently handled: (a) abstraction over sub-strings. e.g., for the spans ``\emph{administration of penicillin}’', ``\emph{administration of aspirin}'', ``\emph{administration of augmentin}'', it could be useful to introduce a shared parent level of ``\emph{administration of antibiotic/drug}''. Our system can currently identify \emph{penicillin}, \emph{augmentin}, \emph{aspirin} as an \emph{antibiiotic/drug}, but cannot handle abstraction over sub-strings. (b) Linking to UMLS currently relies on exact lexical matches, and can be improved.

\section{Ethical Considerations}
We present a system for organizing large result lists into a browsable hierarchy. In general, consuming a hierarchy is more effective than consuming a very long list. However, hierarchies can hide items, especially if the items are misplaced in an unexpected branch---which our system sometimes does (albeit rarely). In situations where consuming the entire information is crucial and the cost of missing an item is prohibitive or dangerous, a flat list would be the safer choice.

\paragraph{Acknowledgements}
This project has received funding from the European Research Council (ERC) under the European Union's Horizon 2020 research and innovation programme, grant agreement No. 802774 (iEXTRACT).

\bibliography{anthology,custom}
\bibliographystyle{acl_natbib}

\end{document}